# Evaluating Large Language Models for Abstract Evaluation Tasks: An Empirical Study

Yinuo Liu[1], Emre Sezgin[1,2], Eric A. Youngstrom[1,2]

*1. Nationwide Children's Hospital, Columbus OH*

*2. Ohio State University, Columbus OH*

Corresponding Author: Dr. Eric A. Youngstrom, Institute of Mental and Behavioral Health Research, Nationwide Children's Hospital, 444 Butterfly Garden Drive, Columbus, OH, 43215, USA. Email: Eric.Youngstrom@nationwidechildrens.org.

## Abstract

**Introduction**: Large language models (LLMs) can process requests and generate texts, but their feasibility for assessing complex academic content needs further investigation. To explore LLM's potential in assisting scientific review, this study examined ChatGPT-5, Gemini-3-Pro, and Claude-Sonnet-4.5's consistency and reliability in evaluating abstracts compared to one another and to human reviewers.

**Methods**: 160 abstracts from a local conference were graded by human reviewers and three LLMs using one rubric. Composite score distributions across three LLMs and fourteen reviewers were examined. Inter-rater reliability was calculated using intraclass correlation coefficients (ICCs) for within-AI reliability and AI-human concordance. Bland-Altman plots were examined for visual agreement patterns and systematic bias.

**Results**: LLMs achieved good-to-excellent agreement with each other (ICCs: 0.59-0.87). ChatGPT and Claude reached moderate agreement with human reviewers on overall quality and content-specific criteria, with ICCs≈.45-.60 for composite, impression, clarity, objective, and results. They exhibited fair agreement on subjective dimensions, with ICC ranging from 0.23-0.38 for impact, engagement, and applicability. Gemini showed fair agreement on half criteria and no reliability on impact and applicability. Three LLMs showed acceptable or negligible mean difference (ChatGPT=0.24, Gemini=0.42, Claude=-0.02) from the human mean composite scores.

**Discussion**: LLMs could process abstracts in batches with moderate agreement with human experts on overall quality and objective criteria. With appropriate process architecture, they can apply a rubric consistently across volumes of abstracts exceeding feasibility for a human rater. However, their weaker performance on subjective dimensions indicates that AI should serve a complementary role in abstract evaluation, while human expertise remains essential.

## Introduction

Peer review is fundamental to evaluations and selections of journals articles, grant applications, and conference presentations, yet this vital process faces two dominant issues: shortage in reviewers and concerns about rater agreement.[1] Reviewing is a time-consuming, voluntary task that requires scholars to spend hours reading material and writing feedback without compensation. Increasing submission volumes of manuscripts and proposals have strained the reviewer pool across scientific fields. As peer-review requests increase, scholars experience growing reviewer fatigue and decline more invitations,[2] leaving journals and conferences reporting greater difficulties at finding qualified evaluators.[3] In addition to the recruitment pressures, judgement reliability has been a persistent issue in scientific review. We would hope that independent reviewers could agree on the content quality and provide similar ratings to the same subjects. However, most researchers have never received formal trainings in peer-review, and relevant resources or curriculum are either limited or difficult to access.[1,4] This lack of standardized trainings contributes to the lack of consensus in content quality evaluation. Studies have documented low-to-fair inter-rater reliability between reviewers for manuscripts and grant proposals,[5–7] raising concerns about inconsistent selection decisions and compromised validity of the peer-review process.

Abstract evaluation exemplifies both problems simultaneously. Reviewers face particularly intense workloads: they must evaluate dozens or hundreds of abstracts across diverse topics at conferences, and this number can increase to thousands for focused topics in systematic reviews.[8] The time-intensive nature of abstract review is magnified when multiple raters score each submission to generate aggregated evaluations, doubling or even tripling the resources required for decision-making. Reliability risks are even heighted for abstract evaluation. Abstracts provide substantially less information than full manuscripts or proposals,[9] making consistent judgments inherently more difficult. Compounding this challenge, reviewers could be assigned abstracts outside their content expertise or evaluate only a subset of submissions rather than the full pool, both common practices in conferences that reduce consistency across reviewers.[10]

Artificial intelligence (AI) offers promising tools to help with these problems. AI technologies have already been applied across various stages of recruiting and selection processes, including writing job advertisements and screening applicant resumes.[11,12] With the emergence of large language models (LLMs)[13]—AI systems trained on text data that could understand and generate human-like languages—there is now a compelling opportunity to use AI to assist in or automate the evaluation of complex scientific material. By integrating LLMs into the evaluation workflow, the research community could alleviate the personnel pressure while devoting greater attention to reliability concerns underlying the traditional review process.

Previous studies have explored and validated AI's potential in academic reviewing. LLMs have showed reasonable accuracy at identifying scientific errors and verifying checklists, supporting their use as reviewing assistants.[14] They could also play the role of reviewers and editors to accelerate the writing of constructive reports or decisions letters, reducing overall reviewing overload.[15] While LLM's actual performance for complete evaluations of manuscripts or proposals needs further investigation, they show good promise in facilitating the peer-review process. In the specific context of abstract evaluation, LLMs have demonstrated strong performance in screening abstracts for systematic reviews and meta-analyses, delivering detailed assessment based on inclusion criteria or evaluating content against appraisal checklists.[16,17] However, when LLM assessments are directly compared with human judgment, important limitations emerge. Shcherbiak et al. reported disappointingly low agreement between AI and human reviewers at scoring conference abstracts,[18] raising questions about the reliability of AI-generated evaluations. Systematic investigation into when and how LLMs can reliably assess scientific material, particularly in high-stake contexts like conferences, is essential to establish the validity and reliability of integrating AI into the research evaluation workflow.

The present study contributes to this discussion through a comprehensive evaluation of LLM performance in abstract scoring. Using abstracts from a local research retreat, we compared three leading AI models—ChatGPT, Gemini, and Claude—across multiple dimensions: scoring consistency, inter-model agreement, and reliability with human expert judgment. This multi-model comparison provides empirical evidence about within-AI reliability and AI-human concordance in a real-world conference context.

## Method

**Sample.** We used research abstracts ($N$=160) from the Abigail Wexner Research Institute (AWRI) Research Retreat 2025 at Nationwide Children's Hospital as grading material. The abstracts were distributed across five research sessions: technology innovation (8.8%), pediatric research (37.5%), animal models (31.3%), intramural funding-related (15.0%), and general topics (43.8%); see Table 1. Authors could choose multiple sessions, with 28.1% of abstracts assigned to more than one category. Each abstract begins with a title and a 50-word summary, followed by a structured abstract body (360-word limit) containing sections on importance, objective, methods, results, and conclusions.

Table 1. Research tracks and major focus

| Session Topic | # of abstracts | Percentage |
|---|---|---|
| Technology Innovation | 14 | 8.8% |
| Pediatric Research | 60 | 37.5% |
| Animal Models | 50 | 31.3% |
| Intramural Funding Related | 24 | 15.0% |
| General Topic | 70 | 43.8% |
| Two or More Sessions | 45 | 28.1% |

**Grading criteria.** One judging rubric was used by human reviewers and LLM agents to evaluate all abstracts (see Table 2). The rubric contained seven individual criteria, each rated on a 5-point Likert scale (1=poor, 5=excellent): impression, clarity, objective, results, impact, engagement, applicability. We also calculated the average of all criterion scores to define a composite score. Together, these eight component scores evaluated the abstract's overall quality (composite, impression), content-specific aspects (clarity, objective, results, impact), and broader relevance on audience interaction and applicability beyond its field (engagement, applicability).

Table 2. Judging rubric used by LLM and human reviewers.

| Abbreviated Name | Full Criterion | Excellent (5) | Good (4) | Moderate (3) | Low (2) | Poor (1) |
|---|---|---|---|---|---|---|
| **Impression** | Overall Impression | Abstract is compelling and reflects high-quality research | Strong overall impression, minor issues | Adequate, meets expectations | Below expectations, significant improvement needed | Does not meet academic or professional standards |
| **Clarity** | Clarity and Organization | Abstract is well-structured, logical, and easy to follow | Mostly clear, minor issues in flow or organization | Adequate clarity but has some organizational issues | Difficult to follow, lacks coherent structure | Unclear and poorly organized |
| **Objective** | Research Question/ Objective | Clearly stated, focused, and significant | Clearly stated but may lack focus or significance | Adequately stated, somewhat vague or broad | Vague or not well explained | Not stated or irrelevant |
| **Results** | Results/ Expected Outcomes | Results are clearly described and support the research objective | Results are described but lack full context | Results or outcomes are vague or too briefly mentioned | Results are unclear or disconnected from objective | No results or expected outcomes presented |
| **Impact** | Significance/ Impact | Clearly explains importance and potential impact of the work | Some explanation of significance, mostly relevant | Limited explanation of impact or relevance | Vague or weak justification for the work's importance | No discussion of significance |
| **Engagement** | Potential for Engagement | High potential for an engaging, interactive, or thought-provoking presentation | Good potential for audience engagement | Some potential for engagement | Low engagement potential | Likely to be unengaging |
| **Applicability** | General Applicability | Topic has wide relevance or practical significance beyond its niche | Some broader applicability or potential impact | Limited applicability or relevance | Relevance is very narrow or unclear | No broader relevance |

**Human reviewers.** Reviewers of abstracts were directors and/or principal investigators at Nationwide Children's Hospital, representing diverse areas of expertise including child mental health, cardiovascular health, biopathology, and other fields. A total of 14 reviewers divided the task of grading 160 abstracts, with each reviewer evaluating a different number of abstracts based on their availability. Human reviewers were masked from AI-generated abstract scores and completed gradings independently. Each abstract was rated by two randomly assigned reviewers.

**Large language models.** We interfaced with ChatGPT-5, Gemini-3-Pro, and Claude-Sonnet-4.5 through API calls to grade abstracts based on the judging rubric. We employed a batch processing strategy, sending ten abstracts per API call to enable cross-comparison of abstracts within each batch, which facilitates the internal calibration of scores and reduces the total number of API requests. Through prompt engineering, we sent the following user prompt to each LLM agent at every API call:

"You are a careful and responsible reviewer grading research abstracts for a conference. The abstracts start with a title and a 50-word description and include the following sections: importance, objective, methods/design, results/findings, conclusions. Here are seven criteria that must be included in the grading: Clarity and Organization, Research Question/Objective, Results/Expected Outcomes, Significance/Impact, General Applicability, Potential for Engagement, Overall Impression. Here are 10 abstracts. Please grade based on the provided rubric."

**Scoring patterns.** We examined scoring patterns by comparing composite score distributions across LLM and human reviewers. We used boxplots to visualize the median, interquartile range, and outliers of composite scores for each rater.

**Inter-rater reliability**. We used intraclass correlation coefficient (ICC) to quantify inter-rater reliability. ICC decomposes the total variance into variance due to true differences between subjects versus variance due to rater differences and measurement error; the higher the proportion attributable to true differences between subjects, the better the rater agreement.[19] ICC values range from 0 to 1, where 0 indicates no agreement and 1 indicates perfect agreement, with thresholds defined as <0.2 poor, 0.2-0.4 fair, 0.4-0.6 moderate, 0.6-0.8 good, 0.8-1.0 excellent.[20]

We calculated ICC for each criterion using two approaches. For agreement among the three LLM agents (ChatGPT-5, Gemini-3-Pro, and Claude-Sonnet-4.5), we used a two-way random effects model with absolute agreement definition and rater average unit, as the same three LLMs evaluated all abstracts. For agreement between human reviewers and each individual LLM, we used a one-way random effects model to account for two random human reviewers grading each abstract.

**Visual agreement patterns**. We constructed Bland-Altman plots[21] to visually inspect the agreement between human reviewers and each LLM on composite scores for 160 abstracts. For each comparison, we plotted the difference between LLM and mean human reviewer scores (y-axis) against the average of the two measurements (x-axis). The plots display the mean difference and 95% limits of agreement (±1.96 *SD*) to identify systematic bias and score-dependent variation in agreement.

## Results

### Scoring patterns

The boxplots of composite scores (see Figure 1) revealed distinct scoring patterns between LLM and human reviewers. The three LLMs, each scoring all 160 abstracts, maintained consistent scoring patterns characterized by narrow interquartile ranges (IQRs) and few outliers. In contrast, the fourteen human reviewers, each grading between 1 and 28 abstracts, exhibited marked heterogeneity in their scoring behaviors, with substantial variations in IQRs, outlier frequencies, and median scores. Among LLM agents, Gemini exhibited lenient scoring with the highest median and widest IQR, while Claude emerged as the most stringent rater with the lowest median and tightest IQR. ChatGPT fell between these two extremes, showing moderate median and variability.

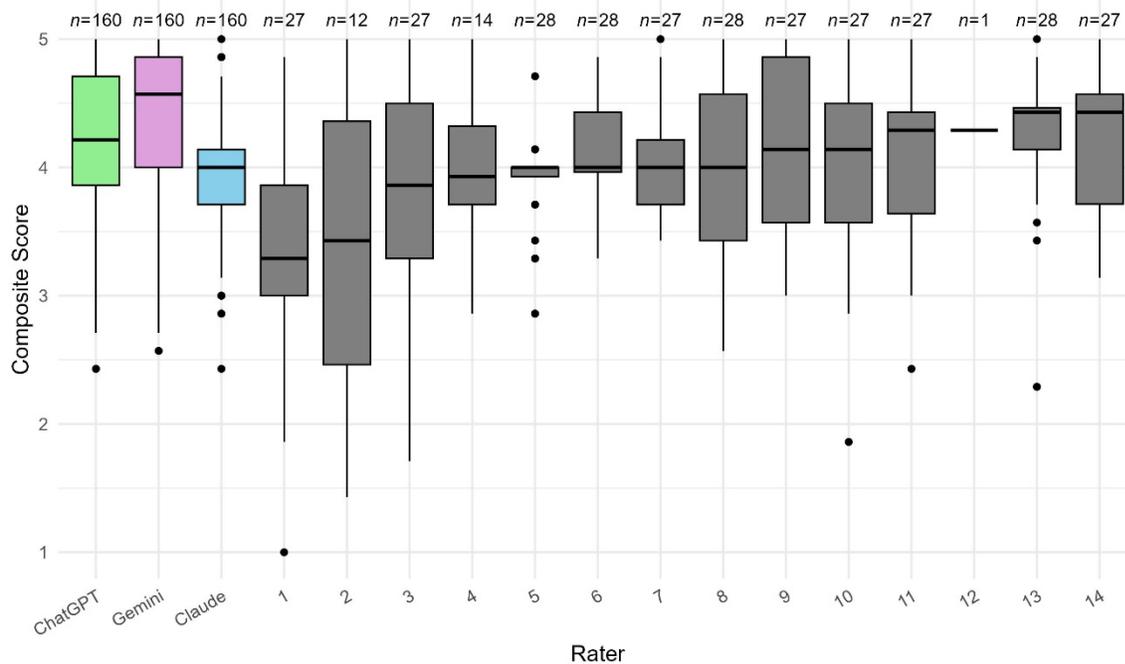

Figure 1. Boxplots of LLM and human reviewers' composite score distribution, averaging across all rubric components

### Inter-rater reliability comparing AI models to each other

ICC analysis between ChatGPT, Gemini, and Claude demonstrated promising inter-rater reliability across AI models (see Figure 2). Two criteria achieved excellent agreement: composite scores (ICC(2,k)=0.80) and results (0.87). Five criteria reached good agreement: impression (0.79), clarity (0.65), impact (0.76), engagement (0.74), and applicability (0.69). Objective (0.59) was the only criterion showing moderate agreement among the LLMs, suggesting greater variability in how these models evaluate the clarity and significance of research questions. Overall, the good-to-excellent agreement across seven out of eight criteria indicates that the three LLMs apply evaluation standards consistently and similarly to each other.

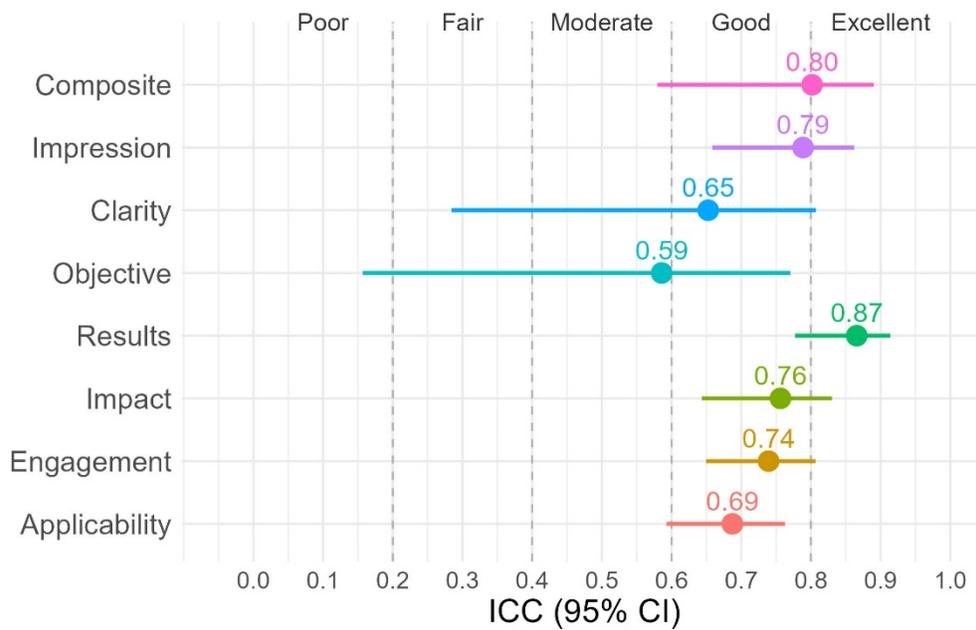

Figure 2. Intraclass correlation (ICC) with 95% CI across each evaluation criterion, comparing ChatGPT, Gemini, and Claude

**Inter-rater reliability comparing human reviewers and AI**

ICC analysis between human reviewers and individual LLM agent revealed mixed results (see Figure 3). In contrast to the good-to-excellent agreement observed among LLM agents, humans and LLMs showed substantially lower inter-rater reliability, with each LLM agent exhibiting a distinct reliability profile.

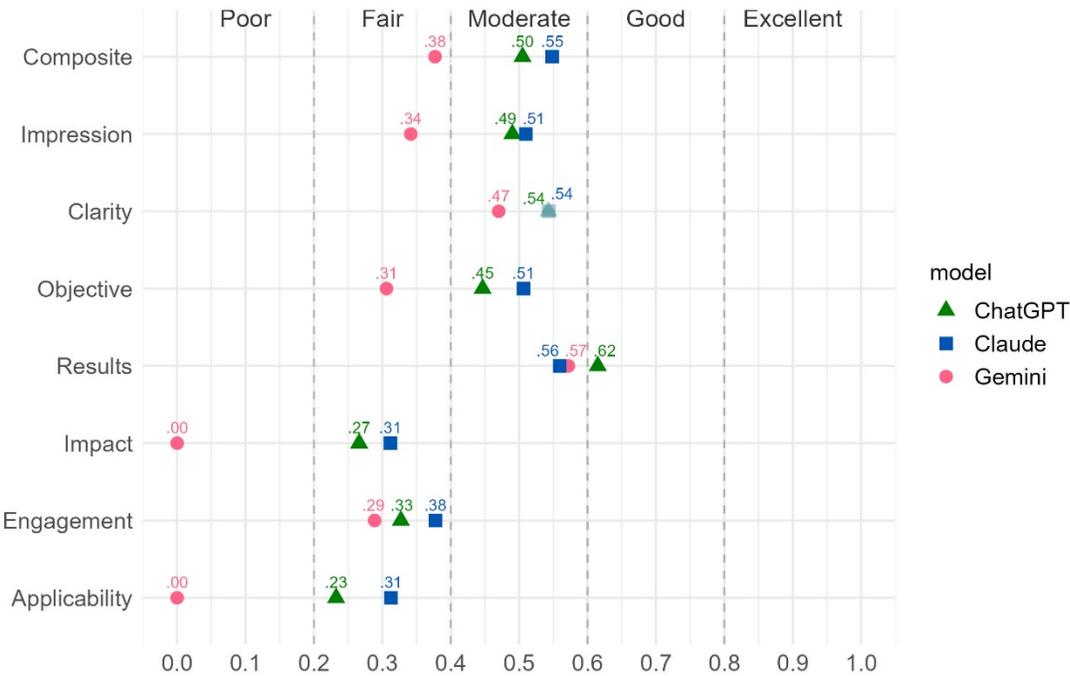

Figure 3. ICC between human reviewers and individual LLM agent

ChatGPT demonstrated fair-to-moderate consistency with human scores: composite (ICC(1,k)=0.50), impression (0.49), clarity (0.54), and objective (0.45) all showed moderate agreement, with results (0.62) achieving borderline good agreement. However, subjective dimensions showed weaker alignment, with impact (0.27), engagement (0.33), and applicability (0.23) demonstrating only fair agreement.

Gemini exhibited the weakest reliability with human reviewers across nearly all criteria: only clarity (0.47) and results (0.57) achieved moderate agreement, while composite (0.38), impression (0.34), objective (0.31), and engagement (0.29) all showed fair agreement. Particularly concerning, Gemini showed no consensus with human reviewers on abstract impact (0.00) and applicability (0.00), suggesting fundamental limitations in this model's ability to assess a work's significance within and beyond its field, similar to how a human reviewer might fare when evaluating a foreign discipline.

Claude demonstrated the strongest alignment with human reviewers across most criteria. Five criteria exhibited moderate agreement: composite (0.55), impression (0.51), clarity (0.54), objective (0.51), and results (0.56). The remaining three criteria showed fair agreement: impact (0.31), engagement (0.38), applicability (0.31). Claude matched ChatGPT's performance in evaluating abstract overall quality and content-specific aspects but proved to be more reliable when assessing subjective dimensions.

**Visual agreement patterns**

The Bland-Altman plots of composite scores visually validated the moderate agreement found between human reviewers and LLM, with additional information on systematic bias and agreement trend (see Figure 4).

A.  B.  C.

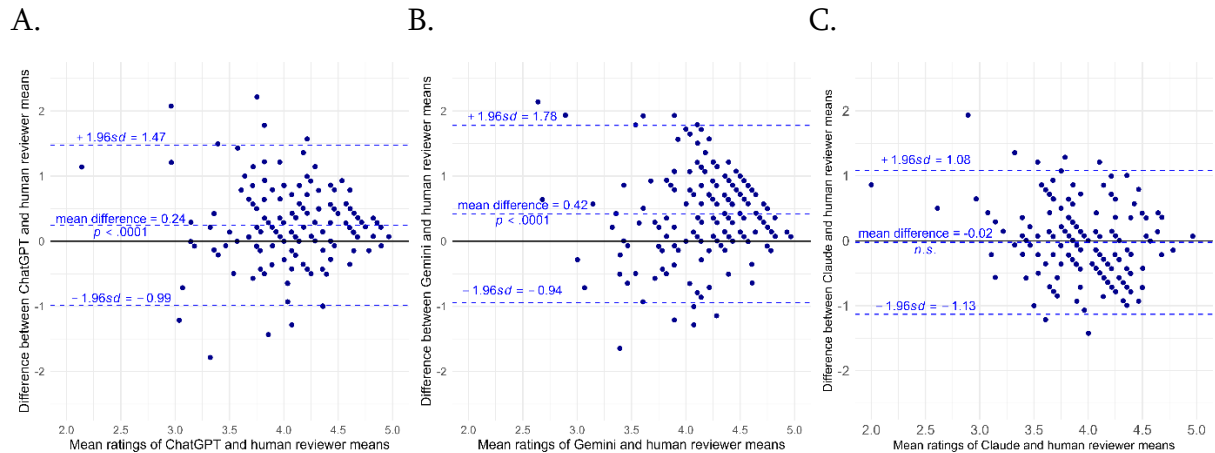

Figure 4. Bland-Altman plots of composite score differences between each LLM and human mean ratings. Panel A: ChatGPT-human comparison. Panel B: Gemini-human comparison. Panel C: Claude-human comparison.

In Figure 4 Panel A, the Bland-Altman plot displays agreement between ChatGPT and human mean ratings. The mean difference of 0.24 (paired t-test, $p$<.0001) indicated a significant but small positive bias, with ChatGPT assigning slightly higher scores on average. About 95% of differences fall between -0.99 and 1.47, reflecting reasonable overall concordance between the two rating methods. Visual inspection of the plot suggested that differences approached zero as average scores increased, particularly for cases with mean ratings above 4.5. Spearman's rank correlation revealed a weak negative association between the absolute differences and mean ratings (Spearman's $\rho$ = –0.16, $p$=0.042), indicating slightly better agreement at higher scores. However, given the small effect size and marginal significance, this trend had limited practical impact and should be interpreted with caution.

Figure 4 Panel B displays the agreement between Gemini and human mean ratings. The two rating methods showed a mean difference of 0.42 (paired $t$-test, $p$<.0001), indicating a systematic positive bias of Gemini scores. The 95% limits of agreement ranged from –0.94 to 1.78, with most observations falling within these limits. There was a weak negative association between the differences and mean ratings (Spearman's $\rho$ = –0.22, $p$<.01), suggesting that the agreement got slightly better when the scores increased. Consistent with the fair reliability observed in the ICC analysis, Gemini exhibited the largest systematic bias and the widest range of disagreement among the three LLMs evaluated.

In Figure 4 Panel C, Claude demonstrated the strongest agreement with human raters, with a near-zero mean difference of -0.02 that was not statistically significant (paired $t$-test, $p$=.5771), corresponding to a negligible systematic bias between the two rating methods. The

95% limits of agreement ranged from -1.13 to 1.08, representing the narrowest range among the three LLMs. Visual inspection suggested a constant spread of differences across the full range of mean ratings, and Spearman's rank correlation confirmed no association between the absolute differences and the mean ratings (Spearman's $\rho$ = 0.03, $p$=0.69), indicating that Claude maintained stable agreement with human raters across all score levels.

### Exploratory cost analysis

Three LLMs respectively graded 160 abstracts within 10 minutes at a cost of less than $1. At this small scale, the three models demonstrated comparable cost efficiency, but actual API cost could increase significantly when grading is scaled to thousands of abstracts or when outputs include detailed qualitative feedback. Generally speaking, using LLM to grade hundreds of scientific abstracts appears efficient and affordable for individuals and institutions with reviewing tasks.

## Discussion

This study evaluated the feasibility of using large language models to review scientific research abstracts from health sciences, bioinformatics, and related topics submitted to a regional conference at a pediatric research hospital. All abstracts were reviewed by at least two human peer reviewers as part of the conference planning process. Three LLMs—ChatGPT-5, Gemini-3-Pro, and Claude-Sonnet-4.5—re-evaluated all abstract using the same scoring rubric and instructions as provided to the human peer reviewers. In terms of inter-rater reliability, ChatGPT and Claude showed moderate agreement with human experts on abstract overall quality and content-specific criteria, with ICCs ranging from 0.5 to 0.6. Both models' reliability on subjective elements dropped to "fair" levels, with impact, engagement, and applicability ratings producing ICCs ranging from 0.2 to 0.3. Gemini performed poorly using human reviewer ratings as the criterion, showing fair agreement on half of the criteria and failing to establish reliability on impact and applicability. Bland-Altman plots further validated the three models' reasonable concordance with human reviewers on abstract overall quality, as all three models showed acceptable or negligible systematic bias compared to human mean ratings, and no strong quality-dependent disagreement pattern was found.

LLMs had practical advantages when serving as additional reviewers for abstract evaluation. First, they provide rapid reviews at a low cost. Human reviewers took over two weeks to complete ratings at considerable expense. A rough estimate suggests 3 minutes per abstract×160 abstracts×2 reviewers≈16 hours of reviewer time, plus the opportunity cost of researchers spending time on reviews rather than research, and the fiscal cost of salary and benefits. Second, simply averaging this LLM review with other ratings will improve overall decision reliability, due to the Spearman-Brown "prophecy" formula (adding another judge of $k$ items has the same effect on reliability as lengthening the test by $k$ items).[22] Third, LLMs could evaluate abstract content independently without direct access to author demographics or prior field knowledge, enhancing their resistance to bias or criterion contamination.[23]

Fourth, LLMs apply one rubric across entire abstract sets and provide uniformly calibrated ratings, mitigating the problem of between-rater variance.[24] When reviewers with different rating standards evaluate separate subsets of abstracts, some applicants are systematically advantaged or disadvantaged based on reviewer assignment rather than content quality. Adding an LLM reviewer thus improves both consistency and equity.

ChatGPT and Claude's moderate agreement with human reviewers confirms that AI-generated scores have reference values for overall quality and objective criteria. This suggests an alternative use case, where LLMs serve as an initial screening mechanism to filter out abstracts with clear deficiencies in structure and data reporting, thus reducing the overall reviewing workload. Such a two-stage review process resembles to triage or selection ratio models often used in admissions or hiring settings.[22,25] However, LLMs' diminished performance in evaluating subjective criteria suggests that they function best as complementary tools in research evaluation, while human expertise remains essential for assessing the subjective dimensions of research quality. Thus, the "extra reviewer" deployment may be preferable to a triage model in settings where subjective criteria are paramount. Rather than replacing human experts, AI reviewers can augment their capabilities, freeing human resource to focus on abstracts requiring nuanced judgment, while enhancing consistency and objective aspects of the review process.

The good-to-excellent consistency between ChatGPT, Gemini, and Claude across evaluation criteria implies that popular LLMs, all built on the transformer structure, respond similarly when given identical user prompts and scientific abstracts. Nevertheless, the remaining variability in abstract scores demonstrates that LLM agents, each trained on distinct datasets and engineered with unique features, develop meaningfully different review styles, evident in their score distributions (Supplemental Figure 1). Claude's stronger alignment with human reviewers and Gemini's notably weaker performance underscore the importance of careful model selection when considering LLMs as tools for scientific abstract review. Conference committees could evaluate models based on their reviewing characteristics (lenient vs. stringent), agreement with human experts, and cost-effectiveness.

Limitations include that sample abstracts predominantly focused on pediatric health research, which does not represent the diversity of topics encountered at national or international conferences. The feasibility of LLM-assisted abstract review in larger-scale, more diverse conference settings, or their performance with humanities and physical science content, requires further investigation. Second, the human reviewers' expertise did not always align closely with the content of their randomly assigned abstracts, and all reviewers used a single "one size fits all" rubric. While peer review at specialized journals typically benefits from better content familiarity and expertise, generalist journals, grant review processes, and conferences with broad scope often encounter similarly wide variation in reviewer-content matching. Under these conditions, the human reviewers achieved only moderate inter-rater reliability—consistent with metrics reported in meta-analyses[6,7]— constraining the upper bound of possible LLM-human concordance.[26] Therefore, the agreement between human reviewers and LLMs should not be interpreted as a gold-standard benchmark for AI's research evaluation capabilities. Rather, this study offers preliminary insights into the potential applications of AI for abstract grading, with findings that warrant further investigation under more controlled conditions. Finally, we only tested three popular AI models—ChatGPT-5, Gemini-3-Pro, and Claude-Sonnet-4.5—which cannot fully represent the variability of rapidly evolving AI models. Our findings should be interpreted cautiously when applied to other models with different architectures or specialized capabilities.

Susskind's distinction between process-oriented versus outcome-oriented AI applications illuminates next steps for AI-assisted research evaluation.[27] Our team adopted a process-oriented approach, providing identical rubrics to AI and human reviewers and evaluating how closely LLM outputs matched human scores. To shift toward the outcome-oriented thinking advocated by Susskind, future work could employ substantially more detailed rubrics with 30 or 40 items that emphasize objective and quantitative criteria where LLMs excel. While such granularity would burden human reviewers, LLMs can consistently apply detailed frameworks at minimal additional cost, as demonstrated by prior works on systematic review and quality coding.[16,17,28] This approach would make AI ratings more complementary to human experts who show competitive advantages at rating subjective dimensions. The psychometric benefits—improved reliability,[22] broader content coverage,[25] and consistent calibration—ultimately serve researchers' need for accurate, unbiased evaluation rather than simply more reviewers.

In conclusion, this study compared popular LLMs with one another and with human reviewers on abstract evaluation by examining score distributions and inter-rater reliability. ChatGPT-5, Gemini-3-Pro, and Claude-Sonnet-4.5 demonstrated scoring efficiency and consistency, with ChatGPT and Claude achieving moderate agreement with human reviewers on overall quality and objective dimensions, but all models showed limited reliability in assessing subjective criteria. Based on these findings, LLMs could serve as complementary tools to augment, rather than replace, human expertise in research evaluation.

Supplemental Figure 1.

Composite score histograms of human mean rating (A), ChatGPT (B), Gemini (C), and Claude (D)

A.

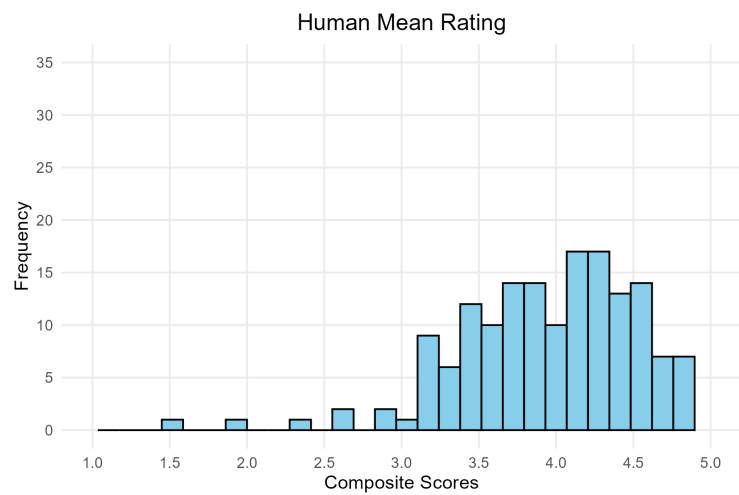

B.

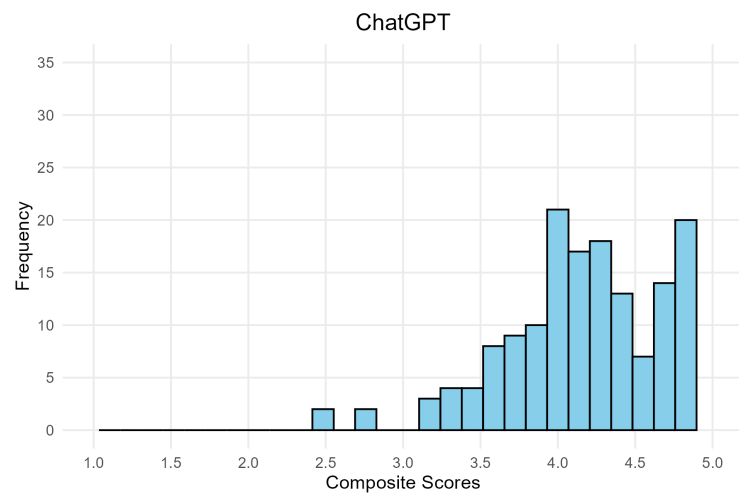

C.

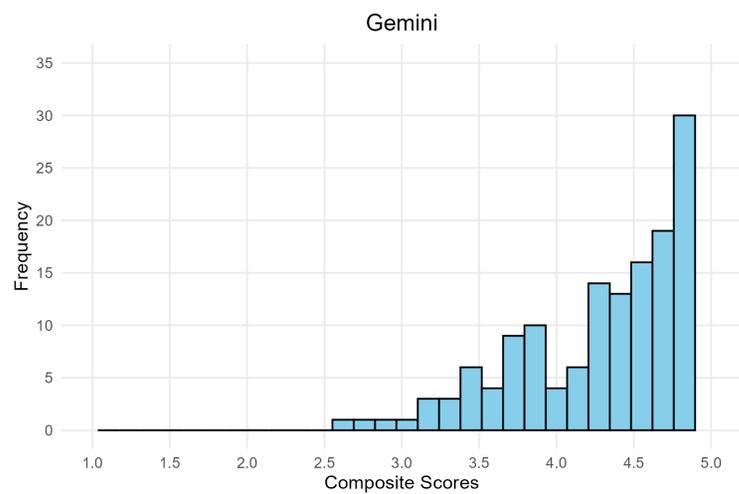

D.

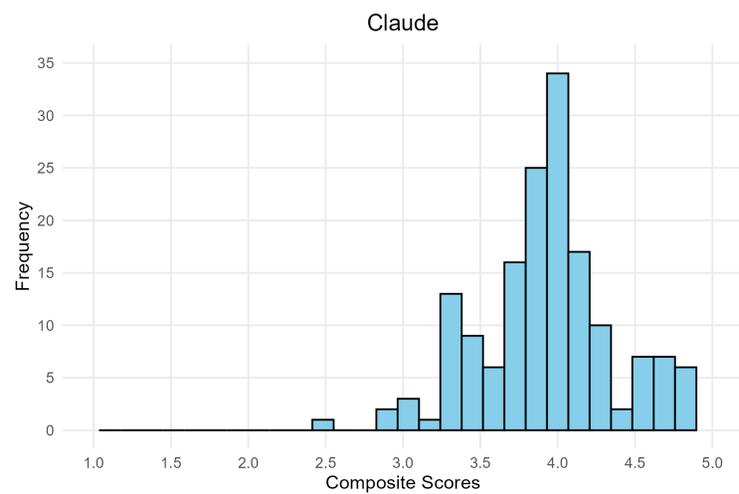